\documentclass[letterpaper, 10 pt, conference]{ieeeconf}  

\IEEEoverridecommandlockouts                              
\usepackage{graphicx} 
\usepackage{url}
\usepackage{xcolor}
\usepackage{amsmath}
\usepackage{multirow}
\title{\LARGE \bf
Dimensional emotion recognition using visual and textual cues
}

\author{Pedro M. Ferreira$^{1}$, Diogo Pernes$^{2}$, Kelwin Fernandes$^{1}$, Ana Rebelo$^{3}$ and Jaime S. Cardoso$^{1}$
\thanks{$^{1}$ Pedro M. Ferreira, Kelwin Fernandes and Jaime S. Cardoso are with Faculty of Engineering of University of Porto (FEUP) and INESC TEC.}%
\thanks{$^{2}$ Diogo Pernes is with Faculty of Science of University of Porto (FCUP) and INESC TEC.}%
\thanks{$^{3}$ Ana Rebelo is with INESC TEC and UCP-ESB.}%
}

\begin{document}

\maketitle
\thispagestyle{empty}
\pagestyle{empty}

\begin{abstract}
This paper addresses the problem of automatic emotion recognition in the scope of the One-Minute Gradual-Emotional Behavior challenge (OMG-Emotion challenge). The underlying objective of the challenge is the automatic estimation of emotion expressions in the two-dimensional emotion representation space (i.e., arousal and valence). The adopted methodology is a weighted ensemble of several models from both video and text modalities. For video-based recognition, two different types of visual cues (i.e., face and facial landmarks) were considered to feed a multi-input deep neural network. Regarding the text modality, a sequential model based on a simple recurrent architecture was implemented. In addition, we also introduce a model based on high-level features in order to embed domain knowledge in the learning process. Experimental results on the OMG-Emotion validation set demonstrate the effectiveness of the implemented ensemble model as it clearly outperforms the current baseline methods. 
%
%
\end{abstract}
\section{INTRODUCTION}
\label{sec:intro}
In psychology, emotion refers to the conscious and subjective experience that is characterized by mental states, biological reactions and psychological or physiologic expressions (i.e., facial expressions) \cite{Li_2011}. Facial expressions are commonly related to affect as they can be defined as the experience of emotion. Together with voice, language, hands and body posture, facial expressions form a fundamental communication system between humans in social contexts.

Automatically perceiving and recognizing human emotion expressions has been one of the key problems in human-computer interaction, with growing application areas including neuromarketing, affect-aware gamification, crowd analytics, biometrics and clinical monitoring \cite{Corneanu_2016}.

Possible sources of input for emotion recognition include different types of signals, such as visual signals (image/video), audio, text and bio-signals. For vision-based emotion recognition, a number of visual cues such as human pose, action and scene context can provide useful information. Nevertheless, facial expression is arguably the most important visual cue for analyzing the underlying human emotions \cite{Zeng_2009,Yu_2015}. 


Most of the available emotion recognition systems are based on the Paul Ekman's categorization scheme \cite{Ekman_1971}, in which emotion expressions are categorized into 6 universal expressions: Happiness, Surprise, Fear, Anger, Sadness, and Disgust. Although these emotional categories are commonly inferred from facial expressions by most people, the way we express ourselves is more gradual and continuous, and hence, most of the times hard to categorize \cite{Barros_2016}. Therefore, dealing with a set of basic emotions could be a serious limitation for many automatic affective systems. Humans usually express themselves differently, sometimes even combining one or more characteristics of the so-called universal emotions. This is somehow embedded in the dimensional emotion representation model, in which emotion expressions are represented in a two-dimensional space, usually arousal and valence \cite{Barros_2016}. This dimensional space represents emotions based on their intensity and nature. For instance, high arousal is usually associated with expressions of high intensity (e.g., excitement) and low arousal with calm and relaxed expressions. High valence is commonly related to positive emotions and low valence to negative emotions. This model has a richer representation of the expressions, without relying on pre-defined categories.

Recent research trends in emotion recognition are based on the dimensional expression representation. It is the example of the One-Minute Gradual-Emotional Behavior challenge (OMG-Emotion challenge) \cite{OMGEmotionChallenge}. The OMG-Emotion competition focuses on long-term emotion recognition in the arousal/valence space. The OMG-Emotion Dataset \cite{OMG_dataset} is composed of 420 relatively long emotion videos with an average length of 1 minute. The videos of the dataset are divided into clips based on utterances, and each utterance is annotated by at least five independent subjects. Each annotator could take into consideration not only the vision and audio information but also the context of each video. That is, each annotator watched the clips of a video in sequence and had to annotate each video using an arousal/valence scale and a categorical emotion based on the universal emotions from Ekman.
%
%

In this paper, an emotion recognition methodology for the OMG-Emotion challenge is presented. The goal is to predict one value of arousal and valence for each video utterance. The implemented methodology is an ensemble of several models from two distinct modalities, namely video and text. More concretely, four different types of models were implemented for the ensemble:
%
%
\begin{itemize}
\item \textit{Face model}: a multi-input deep neural network fed with the extracted faces of the input sequence frames;
\item \textit{Facial landmarks model}: a multi-input deep neural network fed with the facial landmarks of each frame;
\item \textit{Sequential deep text model}: a recurrent deep neural network with an embedding layer initialized with the weights of GloVe \cite{glove};
\item \textit{Feature-engineering text model}: a two-stream multi-layer perceptron fed with \textit{tf-idf} and high-level features.
\end{itemize}

\section{VIDEO-BASED EMOTION RECOGNITION}
\label{sec:video}
For video-based emotion recognition, we designed two different models according to their input nature: 1) the \textit{Face model} that takes directly the face images as input, and 2) the \textit{Facial Landmarks model}, which takes as input 68 key-points located around important facial components (i.e., eyes, nose, and mouth). The purpose of the \textit{face model} is to learn and extract appearance information about facial expressions, which comprises the contour, shape and texture of a face. The \textit{facial landmarks} model explicitly encodes the geometric information about facial expressions.
%
%

\subsection{Pre-processing}
To feed our video-based emotion recognition models, a pre-processing step for face detection and facial landmarks localization is required. To do so, the multi-task CNN face detector \cite{Zhang2016} is first used for face detection and, then, the FAN's state-of-the-art deep learning based face alignment method \cite{Bulat_2017} is used for facial landmarks location. The faces are then normalized, cropped, and resized to $96\times96$ pixels. The facial landmarks coordinates are also normalized by the face image size.
According to Ekman \cite{Ekman_2007}, an expression lasts for $300$ ms to $2$ s. To keep the model simplicity, we extract the face and facial landmarks from a sequence of frames corresponding to $300$ ms. The video sequences of the OMG-Emotion corpus have an average frame rate of approximately 30 f/s, which results in a total of 9 frames as input. Video sequences with higher and lower frame rates are downsampled and upsampled, respectively.

\subsection{Face model}
\begin{figure}[t]
	\centering
	\includegraphics[width=1\columnwidth]{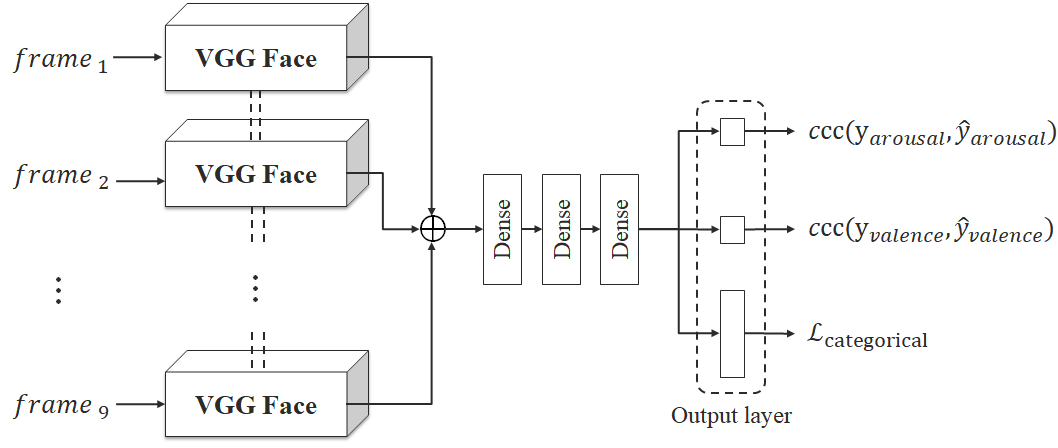}
	\caption{Face model architecture.}
	\label{fig:face_model}
\end{figure}

The implemented face model is an end-to-end multi-input deep neural network. An overview of the network architecture is shown in Figure \ref{fig:face_model}. In particular, the neural network has an input-specific pipe for each frame of the input sequence. Each input-specific pipe is responsible to extract a feature representation of each frame. These feature representations are then merged, followed by a sequence of fully connected layers (or dense layers) with rectified linear units (ReLUs) as non-linearities. The output layer consists of a dense layer with 9 nodes: one for valence, another for arousal, and the remaining for the classification of the 7 categorical emotions (i.e., the 6 universal emotions from Ekman plus the neutral one). While the arousal distribution ranges between $[0,1]$, the distribution of valence varies between $[-1,1]$. Therefore, a sigmoid activation function is used in the arousal node, whereas an hyperbolic tangent is used in the valence one. The neurons of the categorical emotions have a softmax activation function. 

For training the model, the goal is to minimize the following loss function:
\begin{equation}
\label{loss}
\begin{split}
\mathcal{L} =~&~-ccc(\textbf{y}_{arousal}, \hat{\textbf{y}}_{arousal})~-~ \lambda~ccc(\textbf{y}_{valence}, \hat{\textbf{y}}_{valence}) \\ & ~+~\beta~\mathcal{L}_{\textrm{categorical}}(\textbf{y}_{emotion}, \hat{\textbf{y}}_{emotion}),
\end{split}
\end{equation}
where $\lambda,\beta\geq 0$ are the weights that control the interaction of the loss terms. The first two terms of the loss function are defined to maximize the Concordance Correlation Coefficient (CCC) between the model arousal and valence predictions ($\hat{\textbf{y}}_{arousal}$ and $\hat{\textbf{y}}_{valence}$) and their corresponding ground-truth values ($\textbf{y}_{arousal}$ and $\textbf{y}_{valence}$), respectively. The CCC is defined as:
\begin{equation}
ccc(\textbf{y}, \hat{\textbf{y}})~=~\frac{2~\rho(\textbf{y}, \hat{\textbf{y}})~\sigma_{\textbf{y}}~\sigma_{\hat{\textbf{y}}}}{\sigma_{\textbf{y}}^{2}~+~\sigma_{\hat{\textbf{y}}}^{2}~+~(\mu_{\textbf{y}}-\mu_{\hat{\textbf{y}}})^{2}}, 
\label{eq:ccc}
\end{equation}
where $\rho(\textbf{y}, \hat{\textbf{y}})$ is the Pearson's Correlation Coefficient between the ground-truth labels and the model response, $\mu_{\textbf{y}}$ and $\mu_{\hat{\textbf{y}}}$ denote the mean of the ground-truth labels and the model predictions, respectively. $\sigma_{\textbf{y}}^{2}$ and $\sigma_{\hat{\textbf{y}}}^{2}$ are the corresponding variances.

The choice of the CCC as a loss term is motivated by its capability of explicitly demonstrating the model's ability to describe the expressions in a video as a whole, taking into consideration the contextual information \cite{Barros_2016}.

The last loss term, $\mathcal{L}_{\textrm{categorical}}$, trains the model to predict the categorical emotions ($\hat{\textbf{y}}_{emotion}$) given the ground-truth  ($\textbf{y}_{emotion}$) and corresponds to the categorical cross-entropy.
Although the purpose of the OMG-Emotion challenge is not the prediction of the 7 categorical emotions, we use them as an extra supervision layer to regularize the entire learning process.

To work around the problem of training high capacity classifiers in small datasets, such as the one of the OMG-Emotion challenge, the weights of each input-specific stream of the network are shared and initialized with the weights of the VGG-Face network (see Figure \ref{fig:face_model}). The VGG-Face network \cite{Parkhi15} is based on the VGG16 architecture and trained on a very large-scale dataset (2.6M images, 2.6k people) for the task of face recognition. Since the VGG-Face was trained in a similar domain but on a much larger dataset, only the top fully connected layers of our model are fine-tuned during the first training epochs (50 epochs). Afterwards, the whole network is trained, with a smaller learning rate, a few more epochs (15 epochs).

The hyperparameters of the face model, including the weights of the loss function, the $l_{2}$ regularization coefficient, the number of dense layers and neurons per layer, were optimized by means of grid search and cross-validation. 
The best models on the arousal and valence prediction tasks were kept and ensembled by averaging their outputs. Details about the adopted ensemble procedure can be found in section \ref{sec:ensemble}.

\subsection{Facial landmarks model}
The facial landmarks model is topologically identical to the adopted face model, which is also trained to minimize the loss function defined in \eqref{loss}. The facial landmarks model consists of a multi-input neural net with input-specific streams for the facial landmarks coordinates of each frame. In particular, each input-specific pipe consists of a classical neural network with two hidden layers with shared parameters. On top of that, there is also a sequence of dense layers, followed by a final output layer topologically identical to the output layer of the face model (i.e., 9 output nodes with appropriate activation functions).  
%
%
The facial landmarks model is fed with the normalized facial landmarks coordinates (68 key-points $\times$ $x$ and $y$ coordinates) along with a set of temporal and geometric features computed from them. The temporal features attempt to encode how the input facial features changed over time. These features, computed between consecutive frames, include:
\begin{itemize}
\item The velocity of change, computed as the discrete 1st order derivate of the facial landmarks. It measures the rate of change of the per-frame facial features from one frame to the next.
\item The acceleration of change of the per-frame facial landmarks. It is computed as the derivative of the corresponding velocities.
\end{itemize}

The geometric features are computed from the facial landmarks of each frame, individually. The extracted geometric features include:
\begin{itemize}
\item Relative $x$ and $y$ distances between each key-point and the center point of the face;
\item Euclidean distance between each key-point and the center point;
\item Relative angle between each key-point and the center point. The computed angles are corrected by the nose angle offset.
\end{itemize}
These features are then concatenated to form a feature descriptor of each input frame. The hyperparameters of the facial landmarks model
were also optimized by means of grid search and cross-validation.
\section{TEXT-BASED EMOTION RECOGNITION}
\label{sec:text}

\subsection{Sequential deep model}
The adopted sequential model is based on a simple deep recurrent architecture which is also trained to minimize the loss (\ref{loss}). The first layer is a 50-dimensional embedding layer, whose weights were initialized with pre-trained GloVe \cite{glove} word vectors and kept constant during training. The word embeddings are then fed through two cascaded LSTMs \cite{lstm} of size 16. The final output of the recurrent part is applied to a fully connected layer which is structurally identical to the output layer of the face model (9 output nodes with approriate non-linearities applied to each of them). The model was trained using Adam \cite{adam}, with a learning rate of $10^{-3}$, and an $l_2$ regularization coefficient of $10^{-4}$. The relative weights $\lambda$ and $\beta$ of the loss function were cross-validated and the best models on the arousal and valence prediction tasks were kept and ensembled as described in section \ref{sec:ensemble}.
%
%

\subsection{Feature-engineering model}
While end-to-end deep learning strategies are able to achieve state-of-the-art results on large corpus of data, the reduced size of the target dataset difficult the learning of robust models for these tasks. Therefore, we introduce a model based on high-level features that allow to embed domain knowledge.
%
%

In this sense, we extract a Term-Frequency Inverse Document Frequency (\textit{tf-idf}) descriptor from the text and the Part-of-Speech tags of the text. The vocabulary construction includes uni-, bi-, and trigrams. 

Also, we extract high-level features such as:
%
%

\begin{itemize}
\item Sentiment and Subjectivity scores: aggregated polarity, positive/neutral/negative words. We used the standard models from NLTK \cite{bird2009natural} and TextBlob \cite{textblob} to extract these features.
\item Number of tokens in the utterance transcripts.
\item Number of stop-words in the utterance transcripts.
\item Number of swear-words, masked in the dataset using asterisks (*).
\item Number of negations (e.g. don't, not, wouldn't).
\end{itemize}

\begin{figure}[t]
	\centering
	\includegraphics[width=0.6\columnwidth]{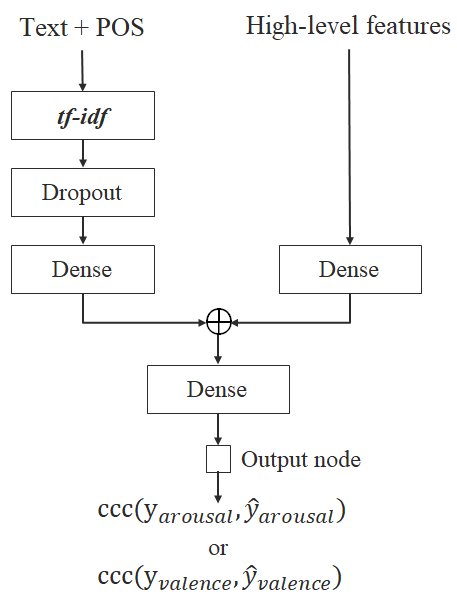}
	\caption{Feature-engineering text model.}
	\label{fig:text_model}
\end{figure}

We aggregated both types of features (i.e. \textit{tf-idf} and \textit{high-level} features) using a two-stream multi-layer perceptron following the architecture illustrated in Figure \ref{fig:text_model}. As done in previous cases, the output activations are the hyperbolic tangent and sigmoid functions to project the network outcome to the target domain. We trained independent models using the CCC objective, defined in \eqref{eq:ccc}.
In order to regularize the learning process of the \textit{tf-idf} stream, which parameters grow linearly with the vocabulary size, we use dropout to simulate the stochastic absence of words in the input text. Also, we consider the test set distribution in the computation of the inverse document frequency terms. This process is known as transductive learning \cite{ifrim2006transductive}.
%
%

\section{Ensemble}
\label{sec:ensemble}
Learning from multi-modal data is a challenging and compelling task, which is usually addressed using at least one of the following strategies: early modality fusion, in which the different modalities are merged in their original space and then fed through the classifier; intermediate fusion, where the modalities are projected and merged in a semantic space and this embedding is then used for classification; late fusion, where independent classifiers for each modality are designed and their predictions are combined via some form of model ensembling.
%
%
For the given dataset, it is a bit impractical to implement an early fusion strategy, given the absence of text transcript for some of the videos in the training set. Moreover, the semantic level of both modalities is also quite different. Thus, if we opted for intermediate fusion, the classifier would be likely to rely mostly on the most represented modality (image data), wasting the useful information provided by the other one (text data). For these reasons, we decided to implement a late fusion procedure, where we compute a weighted average of the predictions of each classifier for each of the two target variables (arousal and valence). The weights for each prediction are given by the CCC score in the validation set of each model and for each variable. This averaging procedure reduces variance in the ensemble classifier, while preserving the relative importance of each individual model.
%
%
\section{EXPERIMENTAL RESULTS}
\label{sec:experiments}
The experimental evaluation of the adopted emotion recognition methodologies was performed using the OMG-Emotion Dataset \cite{OMG_dataset}. This dataset is composed of 420 relatively long emotion videos, collected from a variety of Youtube channels. The videos are divided into clips based on utterances, each of them annotated with arousal and valence values and a categorical label. The dataset as part of the OMG-Emotion competition has a strict evaluation protocol with predefined training, validation, and test sets. In particular, the training, validation and test sets comprise a total of 2442, 617, and 2229 video utterances, respectively. Since we do not have access to the test set labels, the results are reported on the validation set.
%
%

Table \ref{tab:results} compares the performance of the implemented models with the baseline methods of the OMG-Emotion challenge. The results are reported in terms of CCC and mean squared error (MSE) for both arousal and valence target variables.

A first observation, regarding the implemented approaches, is that the best arousal and valence results are achieved by the facial landmarks model and the feature-engineering text model, respectively. However, the most interesting observation is that the adopted multimodal ensemble strategy promotes a significant overall improvement in both arousal and valence results. These results clearly demonstrate the complementarity of both modalities. Finally, it is important to stress that our ensemble model clearly outperforms the four baselines on the validation set of the OMG-Emotion challenge.

\begin{table}
\caption{Results on the OMG-Emotion validation set: (first block) baseline methods, and (second block) implemented methods.}
\centering
\resizebox{0.8\columnwidth}{!}{ 
\begin{tabular}{ccccc}
\hline
\multirow{2}{*}{\textbf{Method}} & \multicolumn{2}{c}{\textbf{Arousal}} & \multicolumn{2}{c}{\textbf{Valence}}\\
& \textbf{CCC} & \textbf{MSE} & \textbf{CCC} & \textbf{MSE} \\
\hline
\textbf{Vision - Face Channel \cite{Barros_2016}} & 0.12 & 0.053 & 0.23 & 0.12\\
\textbf{Audio - Audio Channel \cite{Barros_2016}} & 0.08 & 0.048 & 0.10 & 0.12\\
\textbf{Audio - OpenSmile Features \cite{OMGEmotionChallenge}} & 0.15 & 0.045 & 0.21 & 0.10\\
\textbf{Text \cite{OMGEmotionChallenge}} & 0.05 & 0.062 & 0.20 & 0.12\\
\hline
\textbf{Face model} & 0.18 & 0.067 & 0.32 & 0.16 \\
\textbf{Facial landmarks model} & 0.22 & 0.057 & 0.27 & 0.18 \\
\textbf{Feature-engineering text model} & 0.14 & 0.064 & 0.33 & 0.128\\
\textbf{Sequential text model} & 0.11 & 0.066 & 0.32 & 0.18\\
\textbf{Ensemble} & \textbf{0.23} & \textbf{0.050} & \textbf{0.38} & \textbf{0.12}\\
\hline
\end{tabular}
}
\label{tab:results}
\end{table} 
\section{CONCLUSIONS}
\label{sec:conclusions}
This paper reports our emotion recognition methodology for the OMG-Emotion challenge. The implemented methodology is an ensemble of different models from two distinct modalities, namely video and text. Experiments results demonstrate that our ensemble model clearly outperforms the current baseline of the OMG-Emotion competition.
%
%
%

\addtolength{\textheight}{-12cm}   








{\tiny
\bibliographystyle{ieeetr}
\bibliography{refs}
}

\end{document}